\pdfoutput=1

\documentclass[11pt]{article}

\usepackage[]{ACL2023}

\usepackage{times}
\usepackage{latexsym}

\usepackage[T1]{fontenc}

\usepackage[utf8]{inputenc}

\usepackage{microtype}

\usepackage{inconsolata}

\usepackage{longtable}
\usepackage{supertabular, booktabs}
\usepackage{graphicx}
\usepackage{multirow}
\usepackage{multicol}
\usepackage{diagbox}
\usepackage{slashbox}
\usepackage{balance}
\usepackage{url}
\usepackage{enumitem}
\usepackage{flushend}

\usepackage{siunitx}
\usepackage{geometry}
\usepackage{arydshln}
\usepackage{MnSymbol}
\usepackage{siunitx}

\usepackage{amsmath}
\usepackage{stmaryrd}
\usepackage{algorithm}
\usepackage[noend]{algpseudocode}

\usepackage{subcaption}

\usepackage{xcolor}
\usepackage[normalem]{ulem}
\newcommand\highlight[1][yellow]{%
  \bgroup 
  \markoverwith{\textcolor{#1}{\vrule width.1em height.8em depth.2em}}%
  \ULon 
}

\usepackage{comment}

\title{Does Rationale Quality Matter?\\Enhancing Mental Disorder Detection via Selective Reasoning Distillation}

\author{Hoyun Song \hspace{6mm}
    Huije Lee \hspace{6mm}
    Jisu Shin \hspace{6mm}
    Sukmin Cho \hspace{6mm}\\
    \bf Changgeon Ko \hspace{6mm}
    Jong C. Park$\thanks{\hspace{2mm}Corresponding author}$ \\
    School of Computing \\
    Korea Advanced Institute of Science and Technology (KAIST)\\
    \texttt{\{hysong,huijelee,jisu.shin,nelllpic,pencaty,jongpark\}@kaist.ac.kr}\\
  }

\begin{document}
\maketitle

\begin{abstract}

The detection of mental health problems from social media and the interpretation of these results have been extensively explored.
Research has shown that incorporating clinical symptom information into a model enhances domain expertise, improving its detection and interpretation performance.
While large language models (LLMs) are shown to be effective for generating explanatory rationales in mental health detection, their substantially large parameter size and high computational cost limit their practicality.
Reasoning distillation transfers this ability to smaller language models (SLMs), but inconsistencies in the relevance and domain alignment of LLM-generated rationales pose a challenge.
This paper investigates how rationale quality impacts SLM performance in mental health detection and explanation generation.
We hypothesize that ensuring high-quality and domain-relevant rationales enhances the distillation. 
To this end, we propose a framework that selects rationales based on their alignment with expert clinical reasoning.
Experiments show that our quality-focused approach significantly enhances SLM performance in both mental disorder detection and rationale generation.
This work highlights the importance of rationale quality and offers an insightful framework for knowledge transfer in mental health applications.
The implementation code and dataset are publicly available\footnote{\url{https://github.com/HoyunS/acl25-selective-reasoning-distillation}}.
\end{abstract}

\section{Introduction}

Detecting mental health issues at an early stage is crucial for initiating timely interventions that can significantly improve treatment outcomes.
As online communities continue to grow, researchers in NLP have developed methods to screen user-generated content for signs of depression, anxiety, and other mental illnesses~\cite{jiang2020detection, jiang2021automatic, uban2022multi, ji2022mentalbert, aragon2023disorbert}.
These studies aim to enhance detection accuracy by integrating domain-specific knowledge, such as symptom-related information~\cite{nguyen2022improving, zhang2022symptom, song2023simple, zhang2023phq}.

\begin{figure}
    \centering
    \includegraphics[width=\linewidth]{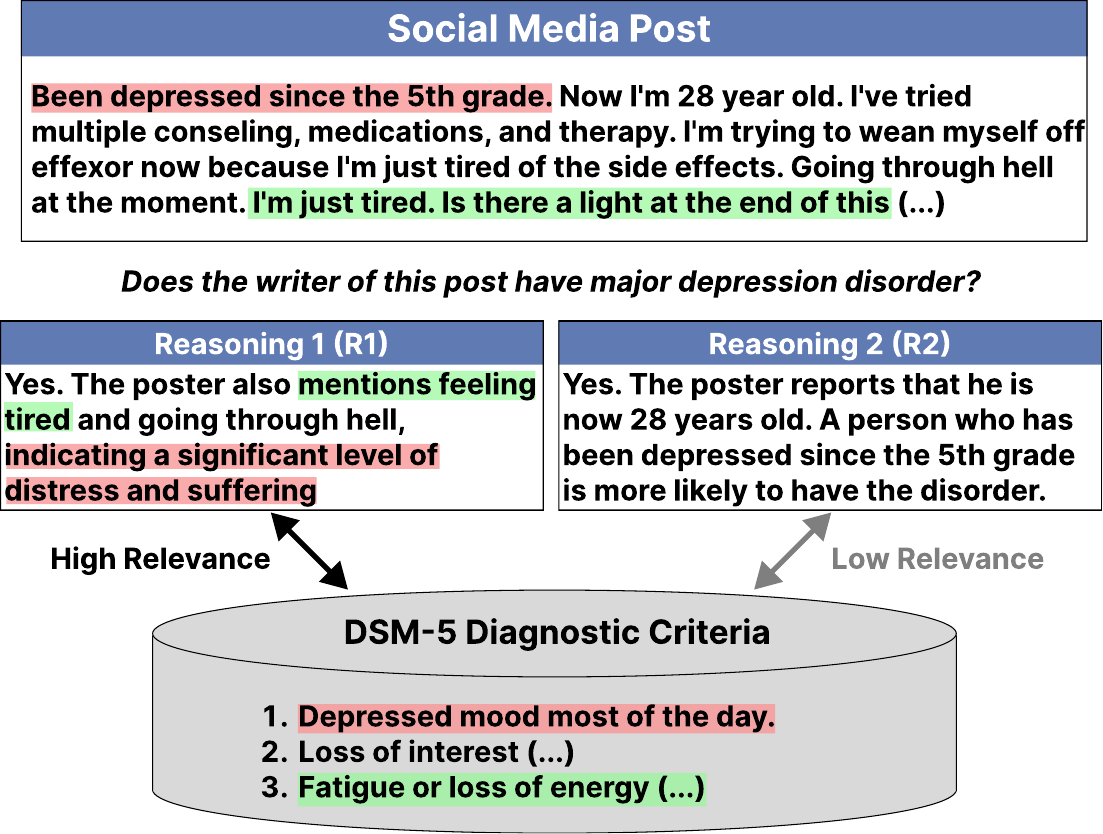}

    \caption{\small Illustration of varying rationale quality. R1 effectively connects the social media post to specific symptoms in the DSM-5 criteria for major depressive disorder, demonstrating high relevance. R2 lacks these connections, showing low relevance. These examples were generated by GPT-3.5.}
    \label{fig:concept_figure}
    \vspace{-0.2in}
\end{figure}


Some studies focus on interpreting detection results to improve diagnostic accuracy and assist mental health professionals in decision-making with clear reasoning~\cite{song2018feature, harrigian2020models, ji2022towards, zanwar2023fuse, malhotra2024xai}.
Large language models (LLMs), with their emerging ability to generate explanatory rationales, have demonstrated promising potential in this area through techniques such as Chain-of-Thought (CoT) prompting~\cite{yang2023towards, wang2024explainable}.
However, this potential comes with a cost, as such reasoning abilities usually require a substantial amount of parameters, which limits their practicality in resource-constrained settings~\cite{wei2022emergent}.

To address these limitations of smaller language models (SLMs) in conducting CoT reasoning, researchers are exploring distillation methods that train these SLMs using teacher-generated rationales~\cite{ho2023large, hsieh2023distilling, magister2023teaching}.
Similarly, \citet{yang2024mentallama} proposed a method for mental health condition classification that leverages the efficiency of SLMs.
They leveraged LLMs to generate explanatory rationale data for fine-tuning an SLM, allowing it to perform similarly to larger models and provide human-like explanations.
Therefore, inspired by these efforts, this study aims to effectively distill the mental health detection ability and rationale generation ability of the teacher model into the student model.

However, despite much advancement in distillation techniques, we found a critical challenge: the inconsistency in the quality of teacher-generated rationales.
We observed that, even with identical inputs, LLMs may produce rationales that vary significantly in their relevance to the specific domain.
For example, as shown in Figure~\ref{fig:concept_figure}, R1 explicitly references established clinical symptoms for major depressive disorder, reflecting how psychiatric professionals reason about diagnoses.
R2, by contrast, is comparatively superficial, lacking references to symptom criteria and failing to incorporate domain-specific insights.
Training on such low-quality rationales may hinder the student model's development of accurate and reliable clinical reasoning, limiting its effectiveness in mental health detection.
Consequently, we hypothesize that, by selectively focusing on high-quality rationales, those strongly aligned with domain-specific knowledge may turn out to be more effective for reasoning distillation in the specific domain.

To implement such a selective approach, we need to define what constitutes a high-quality rationale and establish a method for evaluating it.
Previous research has employed various criteria to assess rationale quality, including fluency, consistency, reliability, and professionality~\cite{jeon2024dual, yang2023towards, yang2024mentallama}.
In this study, we focus on relevance with domain knowledge, a key aspect of professionality. 
Specifically, this criterion refers to the extent to which a rationale is explained based on a sufficient understanding of domain knowledge, as shown in Figure~\ref{fig:concept_figure}.
Our emphasis on domain relevance stems from its close alignment with the reasoning and diagnostic processes of mental health experts, who utilize established clinical criteria~\cite{american2013diagnostic} to identify specific mental disorders.
By prioritizing rationales that reflect this expert reasoning process, we aim to enhance the student model's ability to acquire and apply domain-specific knowledge.

In this paper, we aim to investigate whether selectively distilling high-quality rationales can improve the performance of student models in mental health detection and rationale generation, specifically for major depressive disorder (MDD).
To this end, we propose a framework that includes a process for evaluating and selecting high-quality rationales generated by the teacher model, particularly those that show a strong understanding and integration of domain-relevant knowledge.
This framework will allow us to assess the impact of prioritizing these rationales on the student model's ability to learn and effectively apply clinical reasoning.
Through the experiments, we demonstrated that our quality-evaluation method aligns well with expert reasoning processes, and that the selective distillation method effectively improves the student model's performance in both mental health problem detection and rationale generation.

Our contributions are as follows:
\begin{itemize}[leftmargin=*,topsep=-2px,partopsep=0px]
    \item This is the first study to explicitly investigate the impact of rationale quality on student model performance within the critical context of mental health detection, highlighting the importance of incorporating domain-relevant knowledge into the rationale distillation process for improved mental health detection.
    
    \item We introduce a framework that includes a process for evaluating and selecting high-quality rationales based on domain-specific knowledge, thereby facilitating effective reasoning distillation for mental health detection.
    
    \item Experimental results demonstrate the efficacy of our quality-focused approach, significantly improving student model performance in both mental disorder detection and rationale generation.
    
\end{itemize}

\section{Related Work}

\subsection{Reasoning Distillation from LLMs}
Numerous studies have explored methods for distilling the CoT reasoning capabilities of advanced LLMs into SLMs~\cite{fu2023specializing, ho2023large, hsieh2023distilling, magister2023teaching, wang2023making, dai2024improve}.
These methods typically involve extracting rationales from a teacher model and then fine-tuning a student model on these rationales. 
In the mental health domain, reasoning distillation has also been employed, extracting explanatory rationales from advanced models to address the challenge of limited data for fine-tuning SLMs~\cite{yang2024mentallama}.
This enables the student to elicit reasoning steps and knowledge from the teacher and achieve comparable performance with reduced size and computational cost.

\begin{figure*}
    \centering
    \includegraphics[width=\linewidth]{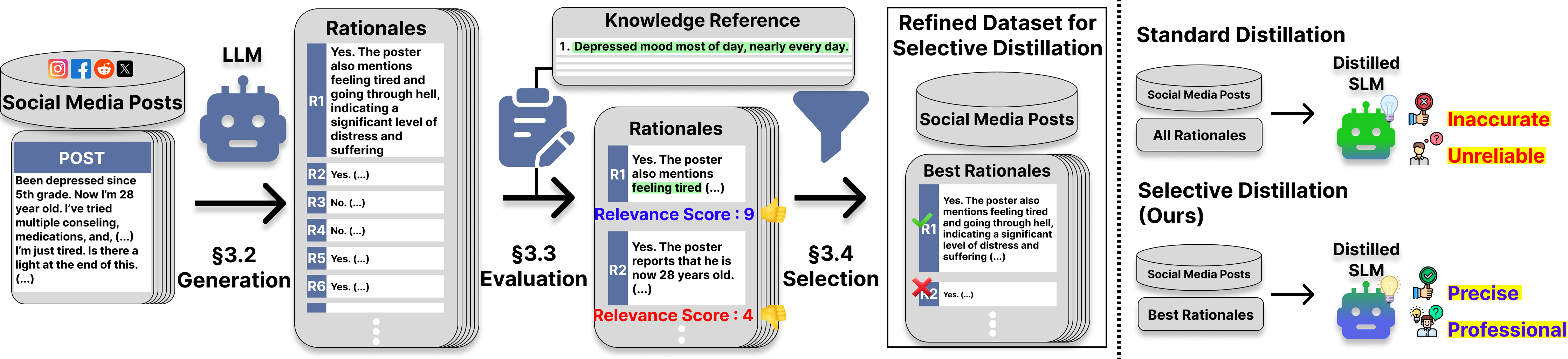}
    \vspace{-0.25in}
    \caption{
    \small Overview of our proposed framework for selective reasoning distillation. Unlike standard reasoning distillation, our framework involves generating various rationales for each post, assessing their quality based on relevance to domain knowledge, and selecting the highest-quality rationale for distillation.
    }
    \vspace{-0.2in}
    \label{fig:method_overview}
\end{figure*}

Since reasoning distillation involves fine-tuning the student model, the quality of the data used for this fine-tuning is important.
Recent studies have demonstrated the effectiveness of using small, high-quality datasets for fine-tuning models, further emphasizing the importance of data quality~\cite{zhou2024lima, xu2024magpie, ye2025limo}.
Researchers also proposed a method to reduce the negative impact from imperfect teacher models, emphasizing the potential for these models to adversely affect students' learning processes~\cite{zhou2024teaching}.
While these studies demonstrate the general importance of data quality, there remains a need to explicitly investigate the impact of selecting high-quality and domain-relevant data in the reasoning distillation process.
This paper focuses on this critical aspect, investigating the impact of rationale quality on SLM performance.


\subsection{Mental Health Detection from Social Media}
Detecting mental disorders through social media involves analyzing a user's posts to determine if they indicate any mental health issues~\cite{yates2017depression, tadesse2019detection, kim2020deep, murarka2020detection, dinu2021automatic, chen2023detection}.
Researchers also investigated interpretable methods that analyze linguistic features, such as emotional expressions~\cite{allen2019convsent, wang2021learning} or specific word choices~\cite{tadesse2019detection, jiang2020detection}, to provide clinically useful insights and explain the model's decisions.
These methods primarily aimed to enhance detection accuracy by screening individuals who may be experiencing mental health issues more effectively.

Given the impressive achievements of LLMs in various domains, recent studies have examined their capability for detecting mental health issues and generating explanatory rationales~\cite{yang2023towards, yang2024mentallama, wang2024explainable, xu2024mental}.
Through CoT prompting, LLMs can include emotional cues~\cite{yang2023towards} and symptom information~\cite{wang2024explainable} while generating detailed, domain-specific explanations.
These studies have also focused on the quality of rationales generated for interpretability, assessing factors such as consistency, reliability, and professionality~\cite{jeon2024dual, yang2024mentallama}.
Inspired by these studies, this research aims to enhance both the performance of mental health problem detection and the rationale quality, thereby improving the clinical applicability of LLMs in mental health detection.

Recent studies have explored integrating standardized diagnostic tools into depression detection models, such as the PHQ-9~\cite{kocalevent2013standardization} and the DSM-5~\cite{american2013diagnostic}.
These approaches aim to improve the accuracy and interpretability of these models by incorporating established clinical criteria and symptom-based assessments~\cite{nguyen2022improving, zhang2022symptom, song2023simple, kang2024cure}.
The utilization of symptom-related information aligns with the diagnostic practices of mental health professionals in real-world clinical settings.
This highlights the importance of incorporating such domain-specific knowledge to provide more clinically relevant and interpretable results.
In line with this emphasis, we hypothesize that high-quality rationales are those that effectively incorporate clinical symptoms, and that fine-tuning a student model with these rationales will enhance both detection performance and rationale quality.

\section{Method}
This section introduces our proposed framework for improving the effectiveness of reasoning distillation by selectively focusing on high-quality rationales for major depressive disorder detection, as illustrated in Figure~\ref{fig:method_overview}.
This framework includes rationale generation, quality evaluation, and quality-based selection.

\subsection{Problem Definition}
We frame our explainable mental health problem detection task similarly to the previous studies~\cite{yang2023towards, yang2024mentallama}.
Suppose that we have a mental health detection dataset \(\mathcal{D} = \{(x_i, y_i)\}_{i=1}^{N}\), where \(x\) represents social media posts and \(y\) their detection labels.
When a post \(x\) is given, we use a teacher model \(\mathcal{T}\) to predict the authors' mental condition \(y'\) and generate rationales \(r\) using CoT prompting \(p\), explaining why each \(x\) is detected as \(y'\).
This process is represented as \((x, y', r) \sim \mathcal{T}(y', r \mid x, p)\).
We determine the detection accuracy by comparing the ground truth label \(y\) and the teacher model's prediction \(y'\).
Through this process, we can collect reasoning training samples: \(\hat{\mathcal{D}} = \{(x_i, y'_i, r_{i})\}_{i=1}^{N}\).
Similar to previous reasoning distillation studies~\cite{magister2023teaching, hsieh2023distilling, ho2023large}, a student model \(\mathcal{S}\) is fine-tuned on \(\hat{\mathcal{D}}\).
We evaluate the clinical interpretability by analyzing the rationales generated by the student models.
Consequently, we aim to enhance SLM performance in both detection accuracy and clinical interpretability by distilling high-quality, domain-relevant rationales from teacher models.

\subsection{Rationale Generation}
First, we utilize a large teacher model to generate explanatory rationale. 
In typical reasoning distillation, when a depression post \(x\) is given, the teacher model is prompted to generate a detection result \(y'\), and an explanatory rationale \(r\).
Unlike typical reasoning distillation, our framework prompts the teacher model to produce multiple diverse rationales for depression posts. 
This creates a pool of candidate rationales that are evaluated for quality, with only the highest-quality rationales selected to fine-tune the student model.
While we utilize temperature adjustments in this study to generate a diverse set of rationales, we note that this is not the only method for achieving this goal.
Details of the prompts for the rationale generation are in Appendix~\ref{appendix:the_prompts_for_rationale_generation}.

\subsection{Rationale Quality Evaluation}
The evaluation of teacher-generated rationale quality, specifically regarding the incorporation of symptom-related information, can be approached through various methods.
Prior work has explored methods such as symptom pattern matching~\cite{nguyen2022improving}, similarity score comparisons~\cite{song2023simple}, and the training of symptom classifiers~\cite{zhang2022symptom}.
Additionally, recent advancements in automated evaluation, especially those utilizing LLMs, have demonstrated potential for achieving human-like assessment~\cite{zheng2023judging}.
Therefore, this study leverages an LLM-based method, selected for its significant performance in previous studies.
However, this does not exclude examining alternative approaches in future research. 

We assessed rationale quality using an LLM-based evaluator, denoted by $\mathcal{E}$. For each post $x_i$ and its corresponding rationale $r_{i,j}$, we compute a quality score \(s_{i,j} = \mathcal{E}(x_i, r_{i,j})\). The evaluator is instructed to assess the alignment between generated rationales and the DSM-5 diagnostic criteria, a standard used by clinical experts in mental disorder diagnosis. 
This assessment is performed by GPT-4o~\cite{achiam2023gpt4} and considers several key factors:
(1) Domain Knowledge: incorporation of clinical symptom-relevant information (i.e., DSM-5 diagnostic criteria); (2) Symptom Recognition: accurate identification of relevant symptoms from the post; and (3) Symptom Relevancy: alignment of the rationale with identified symptom information.
Details of the prompt for the LLM-evaluation are in Appendix~\ref{appendix:the_prompt_for_rationale_quality_evaluation}.


\begin{table*}[ht]
\centering
\scriptsize

\resizebox{\textwidth}{!}{
\begin{tabular}{lcccccccccccc}
\toprule
 \multicolumn{1}{c}{\textbf{Teacher} 
 ($\rightarrow$)} & \multicolumn{6}{c}{\textbf{X (Vanilla Models)}} & \multicolumn{6}{c}{\textbf{GPT-3.5-turbo}} \\
\cmidrule(lr){1-1} \cmidrule(lr){2-7} \cmidrule(lr){8-13}
 \multicolumn{1}{c}{\textbf{Student} ($\rightarrow$)} & \multicolumn{2}{c}{\textbf{Llama-2}} & \multicolumn{2}{c}{\textbf{Llama-3.1}} & \multicolumn{2}{c}{\textbf{Mistral}} & \multicolumn{2}{c}{\textbf{Llama-2}} & \multicolumn{2}{c}{\textbf{Llama-3.1}} & \multicolumn{2}{c}{\textbf{Mistral}} \\
 \cmidrule(lr){1-1} \cmidrule(lr){2-3} \cmidrule(lr){4-5} \cmidrule(lr){6-7} \cmidrule(lr){8-9} \cmidrule(lr){10-11} \cmidrule(lr){12-13}
 \multicolumn{1}{c}{\textbf{Prompt} ($\downarrow$)} & \textbf{Acc.} & \textbf{F1} & \textbf{Acc.} & \textbf{F1} & \textbf{Acc.} & \textbf{F1} & \textbf{Acc.} & \textbf{F1} & \textbf{Acc.} & \textbf{F1} & \textbf{Acc.} & \textbf{F1} \\
\midrule

Std-Cot & 59.24 & 67.24 & 75.82 & 69.36 & 70.51 & 70.88 & 84.38 & 78.15 & 80.49 & 70.75 & 83.01 & 75.69 \\

Std-Cot + Selective & -- & -- & -- & -- & -- & -- & \bf 85.35 & \bf 80.22 & \bf 81.38 & \bf 72.33 & \bf 83.57 & \bf 76.63 \\

\noalign{\vskip 0.1ex}\cdashline{1-13}\noalign{\vskip 0.4ex}

Step-by-Step & 54.34 & 64.84 & 74.34 & 66.17 & 61.69 & 67.60 & 83.98 & 77.71 & 77.56 & 64.43 & \bf 80.53 & \bf 70.82 \\

Step-by-Step + Selective & -- & -- & -- & -- & -- & -- & \bf 85.72 & \bf 80.62 & \bf 83.16 & \bf 75.93 & 79.86 & 69.92 \\

\noalign{\vskip 0.1ex}\cdashline{1-13}\noalign{\vskip 0.4ex}

Emotion & 65.61 & 70.69 & 73.33 & 63.07 & 79.93 & 75.85 & 82.12 & 74.25 & 79.53 & 68.85 & 82.94 & 75.43 \\

Emotion + Selective & -- & -- & -- & -- & -- & -- & \bf 86.05 & \bf 81.27 & \bf 83.12 & \bf 75.84 & \bf 85.94 & \bf 80.69 \\

\noalign{\vskip -0.5ex}
\bottomrule 

\end{tabular}

}

\vspace{-0.15in}

\end{table*}

\begin{table*}[ht]
\centering
\scriptsize

\resizebox{\textwidth}{!}{
\begin{tabular}{lcccccccccccc}
\toprule
 \multicolumn{1}{c}{\textbf{Teacher} ($\rightarrow$)} & \multicolumn{6}{c}{\textbf{GPT-4o}} & \multicolumn{6}{c}{\textbf{Llama-3-70B}} \\
\cmidrule(lr){1-1} \cmidrule(lr){2-7} \cmidrule(lr){8-13}
 \multicolumn{1}{c}{\textbf{Student} ($\rightarrow$)} & \multicolumn{2}{c}{\textbf{Llama-2}} & \multicolumn{2}{c}{\textbf{Llama-3.1}} & \multicolumn{2}{c}{\textbf{Mistral}} & \multicolumn{2}{c}{\textbf{Llama-2}} & \multicolumn{2}{c}{\textbf{Llama-3.1}} & \multicolumn{2}{c}{\textbf{Mistral}} \\
  \cmidrule(lr){1-1} \cmidrule(lr){2-3} \cmidrule(lr){4-5} \cmidrule(lr){6-7} \cmidrule(lr){8-9} \cmidrule(lr){10-11} \cmidrule(lr){12-13}
 \multicolumn{1}{c}{\textbf{Prompt} ($\downarrow$)} & \textbf{Acc.} & \textbf{F1} & \textbf{Acc.} & \textbf{F1} & \textbf{Acc.} & \textbf{F1} & \textbf{Acc.} & \textbf{F1} & \textbf{Acc.} & \textbf{F1} & \textbf{Acc.} & \textbf{F1} \\
\midrule

Std-Cot & 86.24 & 81.29 & 85.57 & 80.04 & 86.57 & 81.68 & 81.86 & 73.29 & 84.76 & 78.58 & 82.38 & 74.17 \\

Std-Cot + Selective & \bf 88.95 & \bf 85.66 & \bf 86.83 & \bf 82.19 & \bf 91.02 & \bf 88.63 & \bf 84.24 & \bf 77.74 & \bf 86.87 & \bf 82.39 & \bf 89.21 & \bf 85.96 \\

\noalign{\vskip 0.1ex}\cdashline{1-13}\noalign{\vskip 0.4ex}

Step-by-Step & 83.53 & 76.58 & 81.64 & 73.10 & 82.27 & 74.19 & 88.91 & 85.62 & 86.65 & 81.89 & 90.62 & \bf 88.36 \\

Step-by-Step + Selective & \bf 89.17 & \bf 85.92 & \bf 85.27 & \bf 79.42 & \bf 90.95 & \bf 88.64 & \bf 89.32 & \bf 86.19 & \bf 86.80 & \bf 82.02 & \bf 90.69 & 88.32 \\

\noalign{\vskip 0.1ex}\cdashline{1-13}\noalign{\vskip 0.4ex}

Emotion & 76.74 & 62.66 & 80.12 & 69.65 & 87.20 & 83.29 & 71.77 & 50.10 & 76.34 & 61.47 & 81.86 & 73.35 \\

Emotion + Selective & \bf 85.76 & \bf 80.49 & \bf 81.90 & \bf 73.25 & \bf 89.35 & \bf 86.39 & \bf 76.93 & \bf 62.80 & \bf 85.05 & \bf 79.15 & \bf 85.98 & \bf 80.95 \\

\noalign{\vskip -0.5ex}
\bottomrule
\end{tabular}

}
\vspace{-0.1in}
\caption{\small Experimental results on the performance of depression detection tasks. Each number represents the accuracy and F1 score of the corresponding student model trained with rationales generated by the corresponding teacher model and CoT prompting strategy.
``+Selective'' indicates that the model was trained using our proposed framework for selectively distilling high-quality rationales.
}

\label{tab:main}
\vspace{-0.1in}

\end{table*}

\subsection{Quality-based Selection}
In the previous steps, we generated a set of \(L\) rationales, denoted by \(\{r_{i,j}\}_{j=1}^L\), for each post $x_i$ using the teacher model.
The best rationale $r_{i,{\text{best}}}$ is selected as:
\begin{equation}
r_{i,\text{best}} = r_{i,j^*} \quad\textnormal{ with } j^* = \underset{j \in \{1,\dots,L\}}{\arg\max}\, s_{i,j}
\end{equation}

Using this process, we construct a refined dataset for selective distillation:
\begin{equation}
\hat{\mathcal{D}}_{\text{SD}} = \{(x_i, y'_i, r_{i, \text{best}})\}_{i=1}^{N}
\end{equation}

Here, only the highest-quality rationale for each post is paired with its corresponding prediction. 
This refined dataset is then used to fine-tune the student model \(\mathcal{S}\), thereby enhancing its ability to produce domain-relevant explanations and improving its performance in mental health detection.

\subsection{Selective Reasoning Distillation}


We fine-tune the student model \(\mathcal{S}\) on the selective distillation dataset \(\hat{\mathcal{D}}_{\textnormal{SD}}\). For each training sample, we concatenate the input post, predicted label, and the selected rationale into a single sequence \(z = [x; y'; r_{\text{best}}]\). The objective loss function is:


\begin{equation}
\mathcal{L}_{\text{SD}} = -\sum_{t=1}^{|z|} \log P(z_t | z_{<t};\theta),
\end{equation}
where \(\theta\) is the set of parameters of \(\mathcal{S}\).

\section{Experiments and Results}
\label{sec4:experiments}

\subsection{Experimental Setup}

\paragraph{Dataset}
To validate the capability of smaller (distilled) models in both mental health problem detection and explanatory rationale generation, we utilize the Reddit\_depression dataset~\cite{song2023simple}.
This dataset comprises Reddit posts from two groups: mental disorder-related subreddits and random subreddits (clean text). Detailed statistics of this dataset are provided in Appendix~\ref{appendix:datasets}.
For evaluating mental disorder detection, we use accuracy and F1 score as the primary metrics. Details of evaluation metrics are described in Appendix~\ref{appendix:details_for_evaluation_metrics}



\paragraph{Models}
For the teacher models, we employed two closed-source LLMs, GPT-3.5~\cite{ouyang2022training} and GPT-4o~\cite{achiam2023gpt4}, and one open-source LLM, Llama-3-70B~\cite{dubey2024llama}.
For the student models, we experiment with Llama-2-7B~\cite{touvron2023llama2}, Llama-3.1-8B~\cite{dubey2024llama}, and Mistral-7B~\cite{jiang2023mistral}.
The details of the model versions are in Appendix~\ref{appendix:versions_of_baseline_models}.

\paragraph{CoT Prompts}
We followed the prompt for generating rationales as suggested by \citet{yang2023towards}.
To generate explanatory rationales, we conducted experiments with the following CoT prompts: (1) \textbf{Std-Cot}~\cite{magister2023teaching, ho2023large}, the standard CoTs distillation method; (2) \textbf{Step-by-Step}~\cite{hsieh2023distilling}, a multi-step reasoning approach; and
(3) \textbf{Emotion-enhanced}~\cite{yang2023towards}, designed to elicit rationales that consider emotional aspects of the input text.
We provide details in Appendix~\ref{appendix:the_prompts_for_rationale_generation}.

\paragraph{Setup}

For rationale generation, we create a pool of 10 rationale candidates for each depression post. 
We set the temperature to 1.0 when using teacher models to generate these candidates.
Other than that, we set the temperature to 0.0 for baseline experiments and during the test phase.
Each request is attempted up to 5 times, with posts being excluded from the dataset if the generation is refused. The evaluator model scores rationales on a scale from 1 to 10.
We employ LoRA~\cite{hu2021lora} for fine-tuning the student SLMs. 
More details of the hyperparameters can be found in Appendix~\ref{appendix:hyperparameters_for_sft_models}.

\paragraph{Human Evaluation Metrics}
We conducted human evaluations to assess the quality of generated rationale, following the guidelines from \citet{yang2024mentallama}.
Three key metrics are employed: (1) \textit{Consistency}, which measures the agreement between the model's diagnosis and the information in the given post; (2) \textit{Reliability}, which evaluates the credibility of the rationale, ensuring it is grounded in facts from the post; and (3) \textit{Professionality}, which evaluates whether the rationale adheres to diagnostic standards. Further details of the evaluation scheme are provided in Appendix~\ref{sec:appendix_human_evaluation_scheme}.

\begin{table}[!tb]
    \centering
    \scriptsize
    \begin{tabular}{l|ccc}
    \hline
         &  \multicolumn{3}{c}{\textbf{Human Evaluation Metrics (0--3$\uparrow$)}} \\
       \textbf{Models}   & \textbf{Consistency} & \textbf{Reliability} & \textbf{Professionality} \\
    \hline
        Llama-2  & {1.88} & {1.62} & 1.56 \\
        \quad+Distillation & \textbf{2.41} & \textbf{2.25} & {1.96} \\
        \quad+Selective & \textbf{2.81} & \textbf{2.69} & \textbf{2.44} \\
    \hline
        Llama-3.1  & 2.15 & 1.93 & 1.70 \\
        \quad+Distillation & 2.32 & 2.22 & 1.85 \\
        \quad+Selective & 2.54 & \textbf{2.62} & \textbf{2.55} \\
     \hline
        Mistral & 1.97 & 1.84 & {1.77} \\
         \quad+Distillation & {2.33} & {2.27} & {1.97} \\
         \quad+Selective & \textbf{2.79} & \textbf{2.66} & \textbf{2.73} \\
     \hline
    \end{tabular}
    \caption{\small Human evaluation results in three metrics. The values that are \textbf{bold} mean the first outperforming groups, determined by Tukey's HSD pairwise test at a significance level of $\alpha=0.05$. 
    } 
    \vspace{-0.1in}
    \label{tab:human_eval}

\end{table}

\subsection{Depression Detection Results}
\label{section4.2}

To comprehensively evaluate the impact of rationale quality on reasoning distillation for mental health detection, we conducted experiments with three distinct teacher models (GPT-3.5, GPT-4o, Llama-3-70B), three different student models (Llama-2, Llama-3.1, Mistral), and three CoT prompting strategies to generate rationales (Std-Cot, Step-by-Step, Emotion-enhanced).
By testing our approach across these various settings, we aimed to demonstrate its robustness and broad applicability.
Table~\ref{tab:main} presents the depression detection performance across various combinations of these models and strategies.

The experimental results show that distilling rationales from the teacher models generally improved the performance of the student models, confirming that domain-specific knowledge can be transferred effectively by reasoning distillation.
Moreover, applying our proposed framework (\textit{+Selective}), which involves selectively distilling high-quality rationales, further enhanced performance across all combinations.
This suggests that focusing on high-quality rationales leads to more effective knowledge transfer.
The consistent improvement observed in the experiments provides compelling evidence that prioritizing rationale quality is essential for maximizing the effectiveness of knowledge distillation in this domain.

\subsection{Rationale Generation Results}
\label{section4.3}

We conducted human evaluations with two domain experts to assess the impact of our quality-focused approach on rationale generation.
The experts hold degrees in psychology and were actively involved in related research. They are familiar with the DSM-5 criteria and other diagnostic tools. 
We randomly sampled 30 examples per model from a total of nine models: three vanilla student models, three models distilled using standard reasoning distillation (\textit{+Distillation}), and three models distilled with selective reasoning distillation (\textit{+Selective}).
All models used the standard CoT prompting strategy with GPT-3.5 as the teacher model.
Each expert independently assessed a total of 270 explanation samples, using three metrics from previous research~\cite{yang2024mentallama}: \textit{Consistency}, \textit{Reliability}, and \textit{Professionality}.
These metrics were rated on a scale of 0 to 3, and we averaged the scores from the two experts~\footnote{Inter-annotator agreement (IAA) was assessed by converting scores to ranks and calculating Cronbach's Alpha.  The resulting score of $\alpha=0.69$ suggests acceptable reliability.}.
The results are presented in Table~\ref{tab:human_eval}.

The results demonstrate that applying our selective distillation method (\textit{+Selective}) consistently improves the quality of the generated rationales, as evidenced by higher scores across all three metrics.
This improvement is particularly pronounced in the \textit{Professionality} metric, indicating that our quality-focused approach effectively enhances the clinical relevance of the explanations.
It suggests that by selectively distilling high-quality rationales, we not only enhance the accuracy of student model predictions but also improve the quality and clinical relevance of their explanations.

\section{Analysis of Quality Evaluation}

\subsection{Evaluation Method Validation}
\label{section5.1}
\begin{table}[!tb]
    \centering
    \scriptsize
    \begin{tabular}{c S[table-format=1.3] S[table-format=1.3] S[table-format=1.3]}
    \toprule
    \textbf{Evaluation Method} & {\textbf{Consistency}} & {\textbf{Reliability}} & {\textbf{Professionality}} \\
    \midrule
    Pattern Matching & 0.075 & -0.001 & 0.223\raisebox{0.3ex}{\tiny **} \\
    BLEU Score & 0.322\raisebox{0.3ex}{\tiny ***} & 0.286\raisebox{0.3ex}{\tiny ***} & 0.257\raisebox{0.3ex}{\tiny ***} \\
    Cosine Similarity & 0.255\raisebox{0.3ex}{\tiny ***} & 0.232\raisebox{0.3ex}{\tiny ***} & 0.180\raisebox{0.3ex}{\tiny *} \\
    BERTScore & 0.374\raisebox{0.3ex}{\tiny ***} & 0.328\raisebox{0.3ex}{\tiny ***} & 0.248\raisebox{0.3ex}{\tiny ***} \\
    LLM-Evaluation & 0.431\raisebox{0.3ex}{\tiny ***} & 0.327\raisebox{0.3ex}{\tiny ***} & 0.565\raisebox{0.3ex}{\tiny ***} \\
    \bottomrule
    \end{tabular}
    \caption{\small Results of comparing different rationale evaluation methods, assessing their correlation with human judgments of \textit{Consistency}, \textit{Reliability}, and \textit{Professionality}. A high correlation indicates a strong alignment with human judgment. (Spearman correlation, *: \(p<.05\), **: \(p<.01\), ***: \(p<.001\))}
    \label{tab:validity}
\end{table}
    
    
    
To assess the validity of our LLM-based evaluation method, we conducted a comparative analysis with other established evaluation metrics.
Our goal was to assess whether our LLM-based approach aligns more closely with human evaluations of rationale quality in comparison to these alternative methods.
Table~\ref{tab:validity} presents the Spearman rank correlation coefficients between the scores generated by each evaluation method and the human evaluation scores across three human evaluation metrics.
The evaluation methods under comparison include symptom pattern matching~\cite{nguyen2022improving}, semantic similarity scores (Cosine Similarity and BERTScore)~\cite{song2023simple}, and BLEU score.

As shown in Table~\ref{tab:validity}, our LLM-based evaluation method exhibits a strong correlation with \textit{Professionality} (0.565, $p<.001$), demonstrating its ability to assess the subtleties of domain-specific knowledge and clinical reasoning, similar to human evaluations.
Significant correlations are also observed for \textit{Consistency} (0.431, \(p<.001\)) and \textit{Reliability} (0.327, \(p<.001\)).
Importantly, these correlations are consistently higher than those achieved against the other automated metrics, suggesting that our LLM-based method aligns more closely with human judgments of rationale quality.

\subsection{Knowledge Reference Comparison}
\label{section5.2}
\label{sec5.2:knowledge_reference_comparison}
\begin{table}[!tb]
    \centering
    \scriptsize
    
    \begin{tabular}{c S[table-format=1.3] S[table-format=1.3] S[table-format=1.3]}
    \toprule
    \textbf{Symptoms (Reference)} & {\textbf{Relevance}}& {\textbf{Corr.w/Human}} & {\textbf{Corr.w/LLM}} \\
    \midrule
    Vocal Nodule& X & 0.172\raisebox{0.3ex}{\tiny *} & 0.307\raisebox{0.3ex}{\tiny ***} \\
    Schizophrenia (DSM-5) & $\triangle$ & 0.383\raisebox{0.3ex}{\tiny ***} & 0.513\raisebox{0.3ex}{\tiny ***} \\
    Anxiety (DSM-5) & $\triangle$ & 0.413\raisebox{0.3ex}{\tiny ***} & 0.563\raisebox{0.3ex}{\tiny ***} \\
    Depression (PHQ-9) & $\bigcirc$ & 0.470\raisebox{0.3ex}{\tiny ***} & 0.716\raisebox{0.3ex}{\tiny ***}\\
    Depression (DSM-5) & $\bigcirc$ & 0.565\raisebox{0.3ex}{\tiny ***} & 1.000 \\
    \bottomrule
    \end{tabular}
    
    \caption{\small Correlation of evaluation scores with human (\textit{Professionality}) and LLM evaluations using varying knowledge sources (X: Irrelevant, $\triangle$: Moderately Relevant, $\bigcirc$: Highly Relevant). Details for each symptom reference are in Appendix~\ref{sec:details_knowledge_references}. (Spearman correlation, *: \(p<.05\), **: \(p<.01\), ***: \(p<.001\))
    }
    \label{tab:reference}
    \vspace{-0.1in}
\end{table}
\begin{figure}[!tb]
    \centering
    \includegraphics[trim={0 1cm 0 0}, width=0.95\linewidth]{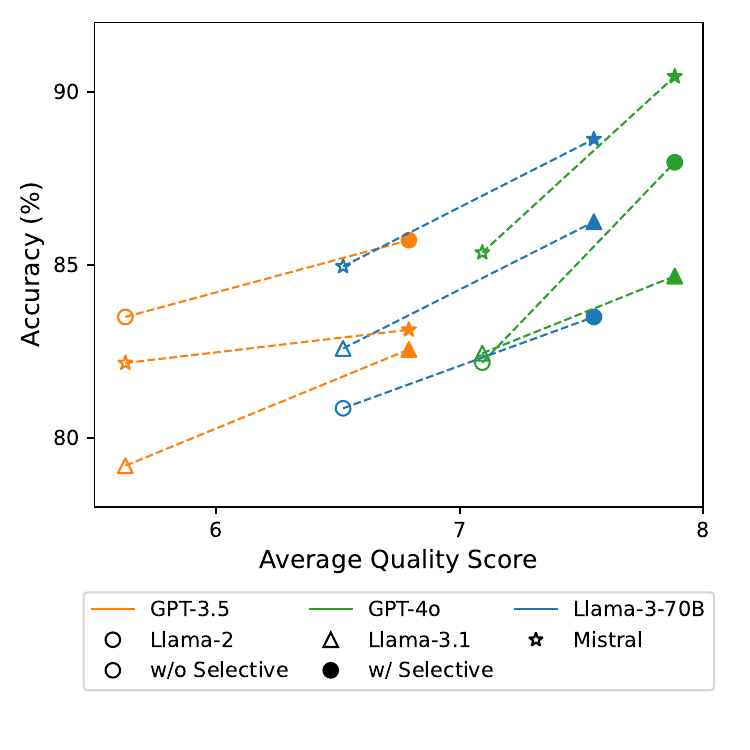}
    \caption{\small Correlation between the quality of teacher-generated rationales and the detection performance of student models. 
    Lines connect the performance of the same student model with and without selective distillation. Markers indicate different student models, while colors indicate different teacher models.}
    \label{fig:quality}
    \vspace{-0.1in}
\end{figure}

\begin{figure*}[!tb]
    \centering
    \includegraphics[trim={1cm 0 1cm 0}, height=4.8cm, width=0.9\linewidth]{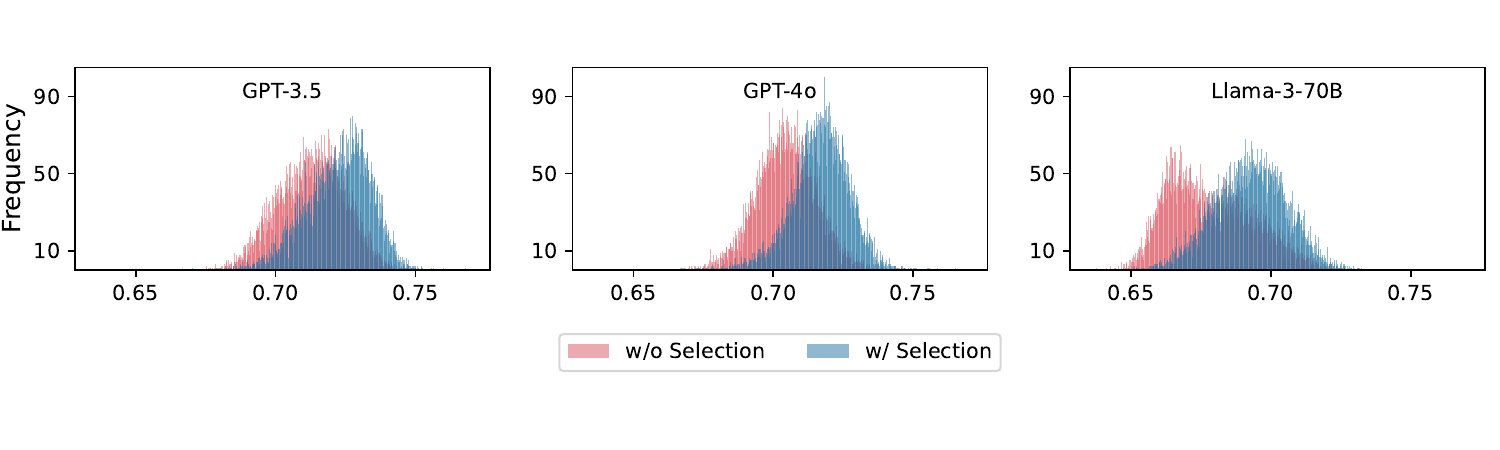}
    \vspace{-0.8cm}
    \caption{\small Distribution of semantic similarity scores between teacher-generated rationales and DSM-5 diagnostic criteria symptom descriptions for depression. We utilized BERTScore to measure the similarity scores and standard CoT prompts to generate rationales. The histograms in each panel, colored in red and blue, represent rationales generated without and with our proposed quality-based selection method, respectively.}
    \label{fig:distribution}
    \vspace{-0.1in}
\end{figure*}
\begin{figure*}[!tb]
    \centering
    \includegraphics[trim={1cm 0.8cm 1cm 0}, height=4cm, width=0.9\linewidth]{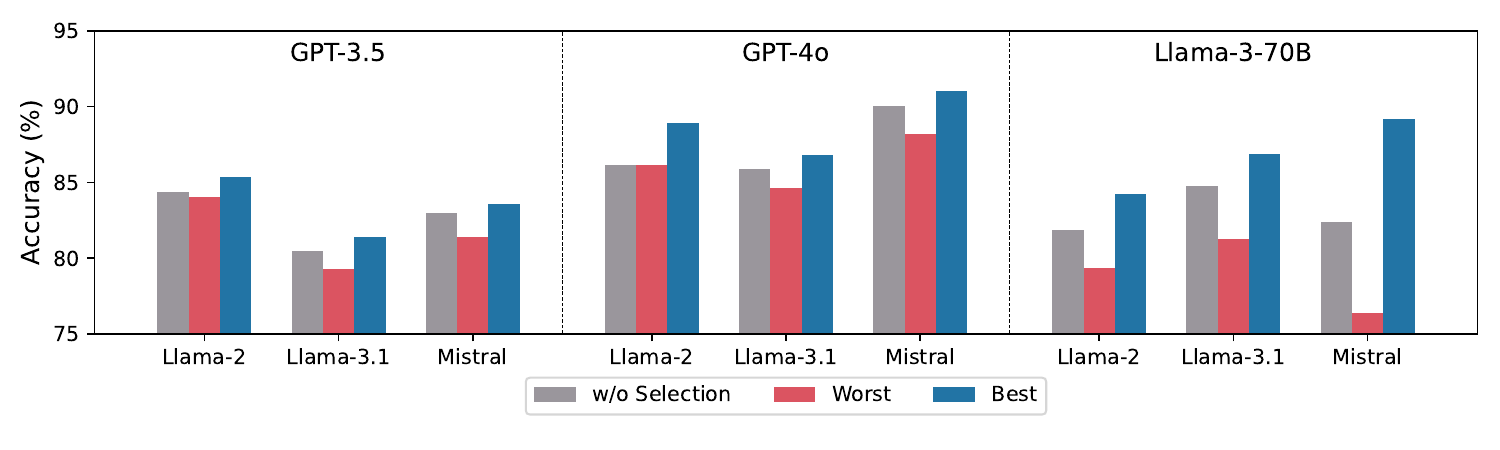}
    \caption{\small Ablation study on different selection criteria. We utilized standard CoT prompts for this experiment. Each bar represents the detection accuracy on the test dataset of the corresponding student model trained with the corresponding teacher.}
    \label{fig:ablation}
    \vspace{-0.1in}
\end{figure*}

We performed a cross-reference analysis to assess how the choice of knowledge source influences our LLM-based evaluation method for major depressive disorder detection.
We used various sources, ranging from irrelevant (e.g., vocal nodule, a physical condition) to highly relevant (e.g., PHQ-9, another depression assessment tool).
We also included references with moderate relevance to mental health, such as those related to other mental disorders (generalized anxiety disorder and schizophrenia from the DSM-5).
Table~\ref{tab:reference} presents the Spearman rank correlations between the evaluation scores obtained using different knowledge references and both human evaluations (\textit{Professionality}) and LLM evaluations.

The results show that, the more relevant the knowledge reference is to mental health, the stronger the correlation with both human and DSM-5 evaluations.
This suggests that our proposed method accurately reflects domain knowledge when assessing rationale quality, highlighting the importance of utilizing a relevant knowledge reference like the DSM-5 for accurate and reliable evaluation.

\subsection{Quality-Performance Correlation}

To investigate how the quality of rationales relates to the performance of student models in detecting mental health issues, we analyzed the correlation between the average quality score of rationales and the accuracy of student models trained on those rationales.
Figure~\ref{fig:quality} illustrates this correlation, plotting the average quality score against the corresponding accuracy.
Each point represents a student model trained with rationales generated by a teacher model. The lines connect points representing the same student model trained with and without our selective distillation method.

Figure~\ref{fig:quality} shows a clear positive trend, indicating that higher rationale quality scores are generally associated with greater accuracy in student models.
While there may be some variation depending on the specific teacher and student models used, the overall pattern suggests that learning with high-quality rationales leads to better performance.
This observation highlights the importance of prioritizing high-quality rationales for effective knowledge distillation in mental health detection.

\section{Analysis of Quality-Based Selection}
\subsection{Clinical Relevance Improvement}

In order to investigate how our quality-based selection approach enhances the clinical relevance of rationales, we compared the distribution of semantic similarity scores between rationales generated by different teacher models and symptom descriptions from the DSM-5 diagnostic criteria for depression. 
We utilized BERTScore~\cite{zhang2019bertscore} to compute the semantic similarity, making it suitable for assessing the alignment between rationales and clinical descriptions.
We visualized these distributions in Figure~\ref{fig:distribution}.
The histograms within each panel show the frequency of different BERTScore values for teacher-generated rationales with and without our quality-based selection.

In most cases, applying quality-based selection results in a noticeable shift of the semantic similarity score distribution towards higher values compared to the \textit{w/o Selection} condition.
This suggests that our method improves the clinical relevance of the rationales by increasing their semantic similarity to the DSM-5 criteria, indicating that they convey more precise and relevant information about the associated symptoms.
This improvement is essential because it ensures that the rationales used for reasoning distillation are clinically meaningful and informative, leading to enhanced student models through fine-tuning with them.

\subsection{Ablation of Rationale Selection}

We conducted an ablation study to analyze the impact of rationale selection by comparing the performance of models trained with different selection strategies: no quality-based selection (\textit{w/o Selection}), selecting highest quality rationales (\textit{Best}), and selecting lowest quality rationales (\textit{Worst}).
Figure~\ref{fig:ablation} illustrates the impact of our quality-based selection method on the detection performance of student models.

As shown in Figure~\ref{fig:ablation}, across all teacher-student model combinations, the \textit{Best} rationales consistently yield the highest accuracy, highlighting the effectiveness of our quality-based selection method.
The \textit{Worst} rationales often result in the lowest accuracy, emphasizing the adverse effect of incorporating low-quality rationales.
These findings demonstrate the importance of prioritizing high-quality rationales for effective knowledge transfer and improved mental health detection.

\section{Conclusion}

This paper empirically investigated how the quality of rationales, especially their relevance to domain knowledge, impacts the rationale distillation process.
We proposed a quality-focused framework that fine-tunes smaller language models with rationales exhibiting high domain relevance, achieved through a process of evaluating and selecting rationales generated by teacher models.
Our experiments demonstrated that our selective reasoning distillation significantly improves both detection accuracy and clinical interpretability in major depressive disorder detection.
These results shed light on the importance of refining distillation data to enhance the model performance and interpretability, especially in mental health applications requiring high-quality domain knowledge.
This work provides a promising direction for enhancing the performance and interpretability of smaller language models in mental health applications. Our selective distillation framework, which prioritizes high-quality, domain-specific knowledge, offers a valuable approach, especially in domains where accurate and reliable explanations are crucial.

\section*{Limitations}

While this study offers valuable insights, it is important to acknowledge that there are some limitations beyond the primary scope of this research.
Each paragraph below discusses these limitations and suggests meaningful directions for future work.

\textbf{Efficient Modeling.}
Our framework incurs significant computational costs due to the necessity of repeated rationale generation and LLM-based quality evaluation. 
These processes, especially with large datasets or complex tasks, can become expensive due to API costs and computational resource requirements.
Although this study prioritizes demonstrating the importance of refining high-quality rationales for enhancing SLM performance, future work should explore more efficient methods. This could involve developing computationally less expensive evaluation metrics or investigating alternative selection strategies that balance quality and efficiency.

\textbf{Collaborative Approach with Experts.}
This study utilizes LLM-generated rationales for distillation, but future work could explore incorporating human-generated rationales to potentially enhance the quality and interpretability of the student models' explanations. 
However, obtaining such data presents practical challenges, including the costs associated with recruiting and compensating experts, as well as the logistical complexities of coordinating data collection and ensuring adherence to privacy and ethical guidelines.
As demonstrated in \S~\ref{section5.1}, our proposed method aligns with human judgment, suggesting its potential as an alternative to direct expert generation of rationales.
Future research could investigate incorporating expert-generated ground truth data into the modeling process, potentially through collaborative approaches or by leveraging expert knowledge to refine evaluation metrics.

\textbf{Clinical Usefulness Evaluation.}
This study incorporates human evaluation (\S~\ref{section4.3}) to assess the quality and interpretability of rationales generated by LLMs, utilizing a method previously established in mental healthcare research~\cite{jeon2024dual, yang2024mentallama}.
However, further research is needed to investigate how to measure both the quantitative and qualitative potential impacts of these rationales in clinical practice.
Evaluating the effectiveness of LLM-generated rationales in real-world clinical settings is crucial to understanding their potential benefits and limitations~\cite{won2025show}.
Future research could investigate more deeply the human-AI interaction regarding how these rationales can assist mental health professionals in diagnosis, treatment planning, and patient communication by conducting user studies with clinicians.

\textbf{Adapting to Different Domains.}
While this study primarily focuses on mental disorder detection, specifically for major depressive disorder, the proposed framework has the potential for broader applicability.
As demonstrated in \S~\ref{section5.2}, incorporating relevant knowledge references can enhance the effectiveness of our framework.
Therefore, by utilizing appropriate knowledge sources, our framework could be adapted for specific domains that require different types of domain knowledge.
Future research should investigate the feasibility and effectiveness of applying this framework to different domains by incorporating relevant knowledge sources and evaluating its performance on diverse tasks.

\textbf{Maintaining Generalizability.}
Generalizability is also an important research topic.
Research on how to maintain performance on general tasks while performing specific tasks is also a very active field of research.
Fine-tuning the student model on a domain-specific dataset may lead to some reduction in generalizability to other tasks.
While this work focuses on optimizing performance for a specific domain, future research could explore strategies to enhance domain-specific performance without sacrificing generalizability.

\textbf{Extending to a Broader Range of LLMs.}
We assessed the robustness and validity of our findings across different architectures by exploring a range of teacher and student models, including both recent and older versions, as well as open-source and closed-source models.
Our experimental results (\S~\ref{section4.2}) demonstrate consistent trends across the models tested, but it is important to recognize that new models are continuously emerging. 
While this work could not encompass all available models, the core principle of training with high-quality, domain-relevant rationales, as demonstrated in this study, remains crucial for achieving optimal performance.
Future research should explore the application of this principle to newly developed models, further investigating how to refine training data to enhance performance across diverse models and domains.

\section*{Ethical Statements}

Our human evaluation was approved by the Institutional Review Board (IRB)\footnote{Approval number: KH2023-166}. All participants were fully informed about the procedure for the evaluation and acknowledged their consent to participate.

To ensure the privacy and anonymity of users, the examples presented in our paper have been paraphrased. Additionally, we adhered strictly to the ethical guidelines proposed for those datasets. The data were used solely for research purposes and were not shared with anyone outside the research team. In our data release for selective reasoning distillation, we will only provide the rationale generated by the model and the index of the original data, excluding any existing dataset content.

While our models showed strong performance in our experiments, it is important to note that these models are not intended to label or stigmatize individuals. Instead, they are designed to serve as tools to enhance public health and provide potential early warnings. We emphasize that any potential mental health-related symptoms and descriptions generated by the models should not be considered definitive diagnoses. For medical diagnosis, the model's output should only serve as supplementary indicators, and consultation with professional psychiatrists or clinical practitioners is essential.


\section*{Acknowledgements}
This work was supported by the National Research Foundation of Korea (NRF) grant funded by the Korea government (MSIT) (No. RS-2023-00208054).

\bibliography{custom}
\bibliographystyle{acl_natbib}

\clearpage
\appendix












\section{Details of Instruction Prompts}

\subsection{Prompts for Rationale Generation}
\label{appendix:the_prompts_for_rationale_generation}

\begin{table}[thb!]
    \small
    \centering
    
    \begin{tabular}{p{0.9\columnwidth}}
        \hline
        \texttt{user:}\\
        Consider this post to answer the question: Is the poster likely to suffer from very severe Major Depressive Disorder? Only return Yes or No, then explain your reasoning. \
        \
        
        
        
        \\
        Post: "\{\textit{Post}\}". \\
        Answer: \\
        \hline
    \end{tabular}
    \caption{The \textbf{Std-Cot} prompt used for rationale generation.}
    \label{tab:std-cot_prompt}
\end{table}

    
        
        
        
        
\begin{table}[thb!]
    \small
    \centering
    \begin{tabular}{p{0.9\columnwidth}}
        \hline
        \texttt{user:}\\
        Consider this post to answer the question: Is the poster likely to suffer from very severe Major Depressive Disorder? Only return Yes or No, then explain your reasoning \textcolor{blue}{\textbf{step by step}}. \
        \
        
        
        
        \\
        Post: "\{\textit{Post}\}". \\
        Answer: \\
        \hline
    \end{tabular}
    \caption{The \textbf{Step-by-Step} prompt used for rationale generation.}
    \label{tab:step-by-step_prompt}
\end{table}

\begin{table}[thb!]
    \small
    \centering
    \begin{tabular}{p{0.9\columnwidth}}
        \hline
        \texttt{user:}\\
        Consider \textcolor{blue}{\textbf{the emotions expressed from}} this post to answer the question: Is the poster likely to suffer from very severe Major Depressive Disorder? Only return Yes or No, then explain your reasoning \textcolor{blue}{\textbf{step by step}}. \
        \
                
        
        
        \\
        Post: "\{\textit{Post}\}". \\
        Answer: \\
        \hline
    \end{tabular}
    \caption{The \textbf{Emotion-enhanced} prompt used for rationale generation.}
    \label{tab:emotion-enhanced_prompt}
\end{table}

        
        
        
        



To evaluate the robustness of our proposed method across different Chain-of-Thought (CoT) prompting strategies, we prepared three distinct CoT prompts. Tables \ref{tab:std-cot_prompt}, \ref{tab:step-by-step_prompt}, and \ref{tab:emotion-enhanced_prompt} present the \textbf{Std-Cot}, \textbf{Step-by-Step}, and \textbf{Emotion-enhanced} prompts, respectively.


\subsection{Prompt for Rationale Evaluation}
\label{appendix:the_prompt_for_rationale_quality_evaluation}

\begin{table}[thb!]
    \small
    \centering
    
    \begin{tabular}{p{0.9\columnwidth}}
        \hline
        \texttt{user:}\\
        You are a clinical psychologist with expertise in diagnosing Major Depressive Disorder (MDD). 
        Your task is to score a reasoning text about potential MDD cases based on symptom identification and logical analysis. Return only a score without rationale.\\
        \\
        Symptom Checklist\\
        \{\textit{Knowledge Reference}\}\\
        \\
        Reasoning Quality Scale\\
        1-2: SEVERELY INADEQUATE\\
        
        No reference to checklist\\
        Unclear or disconnected descriptions\\
        Primarily subjective interpretations\\
        
        3-4: INADEQUATE\\
        
        Mentions symptoms without connecting to checklist\\
        Some relevant points but significant gaps\\
        Overlooks several key symptoms\\
        
        5-6: BASIC\\
        
        Basic references from checklist\\
        Relevant but not insightful observations\\
        Makes logical connections but lacks depth\\
        
        7-8: PROFICIENT\\
        
        Alignment with checklist\\
        Relevant and accurate with minor omissions\\
        Cover key symptoms\\
        
        9-10: EXEMPLARY\\
        
        Clear alignment with checklist\\
        Highly relevant and comprehensive\\
        Consider both key and minor symptoms with clear evidence\\
        \\
        Input Format:""" \\
        Post: [Original text] \\
        Reasoning: [Analysis to evaluate] \\
        """ \\
        
        Output Format:""" \\
        Score: [1-10] \\
        """ \\
        \\
        Post: \{\textit{Post}\} \\
        Reasoning: \{\textit{Rationale}\} \\
        \hline
    \end{tabular}
    \caption{The prompt used for rationale quality evaluation.}
    \label{tab:evaluation_prompt}
\end{table}

To assess how well the rationale considers key factors, we employed GPT-4o as an evaluator using the prompt shown in Table~\ref{tab:evaluation_prompt}. The prompt incorporates the diagnostic criteria of DSM-5 as a knowledge reference, as detailed in Table~\ref{tab:knowledge_reference_dsm5}. Given a post and its corresponding rationale, the evaluator generates scores ranging from 1 to 10.

\section{Details for Experimental Setup}
\label{sec:appendix_experimental_setup}

\subsection{Datasets}
\label{appendix:datasets}
In our experiments, we utilized the Reddit\_depression dataset~\cite{song2023simple}. This dataset comprises Reddit posts annotated for mental disorder detection, focusing on identifying major depressive disorder (MDD). It is divided into training, validation, and test sets with 17,678, 2,696, and 2,696 samples, respectively. Each post is labeled with one of two categories: ``Yes'' (indicating depression) or ``No'' (indicating non-depression).


\subsection{Evaluation Metrics}
\label{appendix:details_for_evaluation_metrics}
To evaluate the performance of the models for mental health problem detection tasks, we use standard classification metrics, including Accuracy (Acc.) and F1-Score (F1). We frame the mental health problem detection task as an explainable mental health analysis, similar to the approach by \citet{yang2024mentallama}. Accordingly, the detection label \(y\) is included in the model's generated output. If a model produces a label that is not part of the predefined detection labels for the task (yes or no), it is considered an unanswered and incorrect response.

\subsection{Versions of Models}
\label{appendix:versions_of_baseline_models}

For the teacher model, 
we selected GPT-3.5 (\textit{gpt-3.5-turbo-0125}) \cite{ouyang2022training}, 
GPT-4o (\textit{gpt-4o-2024-08-06}) \cite{achiam2023gpt4}, and
Llama-3-70B (\textit{meta-llama/Meta-Llama-3-70B-Instruct}) \cite{dubey2024llama} accessed through the DeepInfra API\footnote{\url{https://deepinfra.com/meta-llama/Meta-Llama-3-70B-Instruct}}.
For the student model, we chose Llama-2-7B (\textit{meta-llama/Llama-2-7b-chat-hf}) \cite{touvron2023llama2}, Llama-3.1-8B (\textit{meta-llama/Llama-3.1-8B-Instruct}) \cite{dubey2024llama}, and Mistral-7B (\textit{mistralai/Mistral-7B-Instruct-v0.1}) \cite{jiang2023mistral}. 

\subsection{Hyperparameters for Training and Inference}
\label{appendix:hyperparameters_for_sft_models}

The student models were fine-tuned using LoRA~\cite{hu2021lora}. We used the following parameters: max epoch of 1, batch sizes of 2, gradient accumulation steps of 32, and learning rates of $2e-4$. We used the AdamW optimizer~\cite{loshchilov2017decoupled} with a weight decay of 0.01 and a linear scheduler starting with 50 warmup steps. For the initialization of LoRA weights, we used $r=16$, $\alpha=32$, and a dropout rate of 0.05. All models were fine-tuned on one NVIDIA A100 cluster.

During inference, we set top-p to 0.95, maximum sequence length to 300 tokens, and temperature to 0.0. We utilized vLLM~\cite{kwon2023efficient} to accelerate inference.




\section{Details of Knowledge References}
\label{sec:details_knowledge_references}

\begin{table}[thb!]
    \small
    \centering
    \begin{tabular}{p{0.9\columnwidth}}
        \hline
        \textbf{  DSM-5 diagnostic criteria for MDD}\\
        \hline
        Depressed mood most of the day, nearly every day\\
        Markedly diminished interest or pleasure in all, or almost all, activities most of the day, nearly every day\\
        Insomnia or hypersomnia nearly every day\\
        Significant weight loss when not dieting or weight gain, or decrease or increase in appetite nearly every day\\
        Fatigue or loss of energy nearly every day\\
        Feeling worthlessness or excessive or inappropriate guilt nearly every day\\
        Diminished ability to think or concentrate, or indecisiveness, nearly every day\\
        A slowing down of thought and a reduction of physical movement\\
        Recurrent thoughts of death, recurrent suicidal ideation without a specific plan, or a suicide attempt or a specific plan for committing suicide\\
        \hline
    \end{tabular}
    \caption{DSM-5 diagnostic criteria for major depressive disorder (MDD).}
    \label{tab:knowledge_reference_dsm5}
\end{table}

\begin{table}[thb!]
    \small
    \centering
    \begin{tabular}{p{0.9\columnwidth}}
        \hline
        \textbf{  PHQ-9}\\
        \hline
        Feeling down, depressed, or hopeless.\\
Little interest or pleasure in doing things.\\
Trouble falling or staying asleep, or sleeping too much.\\
Poor appetite or overeating.\\
Feeling tired or having little energy.\\
Feeling bad about yourself - or that you are a failure or have let yourself or your family down.\\
Trouble concentrating on things, such as reading the newspaper or watching television.\\
Moving or speaking so slowly that other people could have noticed.\\
Thoughts that you would be better off dead, or of hurting yourself.\\
        \hline
    \end{tabular}
    \caption{PHQ-9 items assessing depressive symptoms.}
    \label{tab:knowledge_reference_phq9}
\end{table}

\begin{table}[thb!]
    \small
    \centering
    \begin{tabular}{p{0.9\columnwidth}}
        \hline
        \textbf{  DSM-5 diagnostic criteria for GAD}\\
        \hline
        Excessive anxiety and worry, occurring more days than not for at least 6 months, about a number of events or activities.\\
        The individual finds it difficult to control the worry.\\
        The anxiety and worry are associated with irritability.\\
        The anxiety and worry are associated with being easily fatigued.\\
        The anxiety and worry are associated with sleep disturbance (difficulty falling or staying asleep, or restless, unsatisfying sleep).\\
        The anxiety and worry are associated with difficulty concentrating or mind going blank.\\
        The anxiety and worry are associated with muscle tension.\\
        \hline
    \end{tabular}
    \caption{DSM-5 diagnostic criteria for generalized anxiety disorder (GAD).}
    \label{tab:knowledge_reference_anxiety}
\end{table}

\begin{table}[thb!]
    \small
    \centering
    \begin{tabular}{p{0.9\columnwidth}}
        \hline
        \textbf{  DSM-5 diagnostic criteria for schizophrenia}\\
        \hline
        The presence of one (or more) delusions with a duration of 1 month or longer.\\
            Criterion A for schizophrenia has never been met. Note: Hallucinations, if present, are not prominent and are related to the delusional theme (e.g., the sensation of being infested with insects associated with delusions of infestation).\\
            Apart from the impact of the delusions, or its ramifications, functioning is not markedly impaired, and behavior is not obviously bizarre or odd.\\
            If manic or major depressive episodes have occurred, these have been brief relative to the duration of the delusional periods.\\
            The disturbance is not attributable to the physiological effects of a substance or another medical condition and is not better explained by another mental disorder, such as body dysmorphic disorder or obsessive-compulsive disorder.\\

        \hline
    \end{tabular}
    \caption{DSM-5 diagnostic criteria for schizophrenia.}
    \label{tab:knowledge_reference_schizophrenia}
\end{table}

\begin{table}[thb!]
    \small
    \centering
    \begin{tabular}{p{0.9\columnwidth}}
        \hline
        \textbf{  Diagnostic features of vocal nodules}\\
        \hline
        Experiences persistent hoarseness or a raspy voice, especially after speaking for long periods.\\
            Notices frequent vocal fatigue or difficulty projecting the voice.\\
            Feels a sensation of strain or pain in the throat when speaking or singing.\\
            Has a reduced vocal range, particularly in higher or lower pitches.\\
            Experiences frequent throat clearing or the feeling of something stuck in the throat.\\
            Has a history of overusing the voice, such as yelling, shouting, or excessive speaking.\\
            Feels dryness or irritation in the throat despite staying hydrated.\\

        \hline
    \end{tabular}
    \caption{Diagnostic features of vocal nodules.}
    \label{tab:knowledge_reference_vocal_nodule}
\end{table}

We present the knowledge references used in \S\ref{sec5.2:knowledge_reference_comparison}. Tables \ref{tab:knowledge_reference_dsm5}, \ref{tab:knowledge_reference_anxiety}, and \ref{tab:knowledge_reference_schizophrenia} present the DSM-5 diagnostic criteria for major depressive disorder (MDD), generalized anxiety disorder (GAD), and schizophrenia, respectively~\cite{american2013diagnostic}. Table~\ref{tab:knowledge_reference_phq9} presents the PHQ-9 items~\cite{kocalevent2013standardization}, a self-administered screening tool used to diagnose and assess a severity of depression. Table~\ref{tab:knowledge_reference_vocal_nodule} describes the diagnostic features of vocal nodules from Wikipedia\footnote{\url{https://en.wikipedia.org/wiki/Vocal_cord_nodule}}, which serves as an irrelevant source of knowledge compared to MDD diagnosis.

\section{Human Evaluation Scheme}
\label{sec:appendix_human_evaluation_scheme}

We recruited two domain experts specializing in clinical psychology, providing each with a compensation of \$100 for their evaluation.
The evaluation criteria proposed by \citet{yang2024mentallama} were as follows:

\textbf{Consistency}: Evaluates whether the rationale is consistent with the detection result determined by the teacher model for the given post and if the rationale sufficiently supports the detection decision.
\begin{itemize}
    \item 0: The detection result and the explanation do not match.
    \item 1: The detection result and the explanation match, but the explanation is difficult to read and contains serious errors.
    \item 2: The detection result and the explanation match. The explanation is mostly consistent and readable, with a few minor errors.
    \item 3: The detection result and the explanation match perfectly. The explanation is natural, consistent, and error-free.
\end{itemize}

\textbf{Reliability}: Assesses the trustworthiness of the generated rationale, ensuring that it is fact-based and reliable.
\begin{itemize}
    \item 0: Completely untrustworthy and contains false information (e.g., non-existent symptoms).
    \item 1: Partially trustworthy but includes explanations not based on facts.
    \item 2: Mostly trustworthy but contains minor misinformation or incorrect explanations.
    \item 3: Completely trustworthy.
\end{itemize}

\textbf{Professionality}: Evaluates if the rationale adheres to diagnostic standards from clinical descriptions. Annotators used information from PHQ-9 and DSM-5 to determine symptoms and sorted them based on their own knowledge.
\begin{itemize}
    \item 0: The explanation lacks any evidence or omits critical symptoms.
    \item 1: The explanation provides a few supportive pieces of evidence but omits more critical symptoms.
    \item 2: The explanation includes several supportive pieces of evidence but omits some minor symptoms.
    \item 3: The explanation includes all related supportive pieces of evidence in the post.
\end{itemize}

\section{Case Study}
\label{sec:appendix_case_study}
\begin{table*}[t]
\small
    \resizebox{\textwidth}{!}{
        \begin{tabular}{c|p{0.93\textwidth}}
            \hline
            \multicolumn{2}{c}{
                \begin{tabular}[c]{@{}c@{}}
                    Post
                \end{tabular}
                }   \\
            \hline
            \multicolumn{2}{c}{
                \begin{tabular}[c]{@{}p{1\linewidth}@{}}
                    No one seems to care about me.  My crush, who dumped me, is texting me again but mentions other guys. My family problems make me want to be alone for the rest of my life and just cry. I had high hopes for 2021, but I'm still not expecting anything good to happen.  Normally my parents are disappointed in me; I was the dumbest kid in the class. I never expect to accomplish anything in life; I fully expect to die alone. I have zero energy for anything; I just lie in bed and want to cry nonstop. I still haven't forgotten her. I already know everything sucks for me, and that I'll never find happiness. I just hope the world ends soon.
                \end{tabular}
            }   \\
            \hline
            Method  & \multicolumn{1}{c}{Responses}    \\
                
            \hline
            +Distillation & \textbf{Yes}. 
            \newline 1. \textcolor[rgb]{0.8, 0.8, 0.0}{$\triangle$}\textbf{Self-Perception and Emotional Confusion}: The poster describes feeling disconnected from their emotions and struggling to understand their thoughts and feelings. This level of self-awareness and emotional confusion is often associated with Major Depressive Disorder (MDD).
            \newline 2. \textcolor[rgb]{0.8, 0.8, 0.0}{$\triangle$}\textbf{Doubt and Self-Doubt}: The poster expresses doubt about their ability to talk, reason, and empathize with others, which can be indicative of low self-esteem and self-doubt, common in depression.
            \newline 3. \textcolor[rgb]{0.8, 0.8, 0.0}{$\triangle$}\textbf{Anxiety and Fear}: The mention of a ``sinking feeling of dreadful awfulness and revolting grossness'' suggests a high level of anxiety and fear, which are common symptoms of depression.
            \newline 4. \textcolor[rgb]{0.0, 0.5, 0.0}{$\checkmark$}\textbf{Lack of Clarity and Confusion}: The poster expresses a lack of understanding about their emotions and feelings, which can be a sign of depression.
            \newline 5. \textcolor[rgb]{0.7, 0.0, 0.0}{X}\textbf{Seeking Help and Uncertainty}: The poster is seeking help and acknowledges that they may be experiencing something wrong, indicating a level of self-awareness and willingness to seek help, which are positive signs.\\
            \hline
            +Selective & \textbf{Yes}. 
            \newline 1. \textcolor[rgb]{0.0, 0.5, 0.0}{$\checkmark$}\textbf{Expression of Hopelessness}: The poster expresses a deep sense of hopelessness and a belief that nothing good will happen in their life, which is a key symptom of Major Depressive Disorder (MDD). 
            \newline 2. \textcolor[rgb]{0.0, 0.5, 0.0}{$\checkmark$}\textbf{Lack of Energy and Motivation}: The poster mentions having ``0 energy for anything'' and wanting to cry nonstop, indicating a significant lack of motivation and energy, which are common symptoms of depression.
            \newline 3. \textcolor[rgb]{0.8, 0.8, 0.0}{$\triangle$}\textbf{Social Isolation}: The poster feels like no one cares about them and describes a lack of support from family and friends, which can exacerbate depressive symptoms.
            \newline 4. \textcolor[rgb]{0.0, 0.5, 0.0}{$\checkmark$}\textbf{Negative Self-Perception}: The poster describes themselves as ``the dumbest kid in the class'' and expresses a belief that they will never accomplish anything in life, which reflects a negative self-image and low self-esteem. 
            \newline 5. \textcolor[rgb]{0.0, 0.5, 0.0}{$\checkmark$}\textbf{Suicidal Ideation}: The poster mentions hoping for the world to end faster so they could be ``put down,'' which is a serious indicator of suicidal ideation and a critical\\
            \hline
        \end{tabular}
}
\caption{ \small
Example rationales generated by distilled student models assessing a social media post for detecting major depressive disorder and generating a rationale. For these examples, we utilized GPT-4o as a teacher model and Llama-2 as a student model. The notation \textcolor[rgb]{0.0, 0.5, 0.0}{$\checkmark$} indicates the high relevance with clinical symptoms, referring to proper symptoms in the post. 
\textcolor[rgb]{0.8, 0.8, 0.0}{$\triangle$} indicates a sign of depression but not related to diagnostic criteria. \textcolor[rgb]{0.7, 0.0, 0.0}{X} indicates not proper reasoning. To ensure the privacy and anonymity of users, the example post presented in this table have been paraphrased.
}
\label{tab:data:case_study1}
\end{table*}

This section presents a case study to illustrate the practical application of our framework for selective reasoning distillation. 
We analyze the performance of student models trained with (\textit{+Selective}) and without (\textit{+Distillation}) our quality-based selection framework, focusing on their ability to detect major depressive disorder (MDD) and generate clinically relevant rationales. 
Tables \ref{tab:data:case_study1} and \ref{tab:data:case_study2} present the example responses generated by distilled Llama-2 and Mistral student models for the corresponding social media posts, respectively, after being trained on rationales produced by the GPT-4o and Llama-3-70B teacher models, respectively.

In both tables, the \textit{+Selective} models generate rationales that are more focused, relevant, and clinically informative than the \textit{+Distillation} models, demonstrating the effectiveness of our framework in enhancing the quality and interpretability of rationales generated by distilled student models.
While the specific writing styles of the rationales vary depending on the student and teacher models used, applying our quality-based selection framework generally enhances interpretability.
These case studies highlight the practical benefits of our framework for selective reasoning distillation, showing that by selectively distilling high-quality rationales, we can enhance the ability of student models to generate clinically relevant and interpretable explanations for mental health conditions.

\begin{table}[ht!]
    \centering
    \small
    \begin{tabular}{llcc}
        \toprule
        \multicolumn{1}{c}{\textbf{Category}} & \multicolumn{1}{c}{\textbf{Model}} & \textbf{Acc.} & \textbf{F1} \\
        \midrule
        \multirow{3}{*}{Teacher models} & GPT-3.5 & 80.71 & 74.14 \\
                       & GPT-4o & 86.92 & 83.94 \\
                       & Llama-3-70B & 86.86 & 83.79 \\
        \midrule
        \multirow{2}{*}{Latest models} & MentalLLaMA-13B & 74.60 & 70.70 \\
                      & MentalLLaMA-7B & 79.57 & 78.57 \\
        \midrule
        \multirow{3}{*}{Vanilla student} & Llama-2-7B & 59.24 & 67.24 \\
                        & Llama-3.1-8B & 75.82 & 69.36 \\
                        & Mistral-7B & 70.51 & 70.88 \\
        \midrule
        \multirow{3}{*}{Ours} & Llama-2-7B + ours & \underline{88.95} & \underline{85.66} \\
             & Llama-3.1-8B + ours & 86.83 & 82.19 \\
             & Mistral-7B + ours & \bf 91.02 & \bf 88.63 \\
        \bottomrule
    \end{tabular}
    \caption{\small Results of the detection performance comparing our method with recent LLMs.}
    \label{tab:sota_performance}
\end{table}

\begin{table*}[!tb]
\centering
\small
\begin{tabular}{ll cc cc cc}
\toprule
\multicolumn{1}{c}{\multirow{2}{*}{\textbf{Category}}} & \multicolumn{1}{c}{\multirow{2}{*}{\textbf{Model}}} & \multicolumn{2}{c}{\textbf{DR}} & \multicolumn{2}{c}{\textbf{Dreddit}} & \multicolumn{2}{c}{\textbf{Reddit\_anxiety}} \\
\cmidrule(lr){3-4} \cmidrule(lr){5-6} \cmidrule(lr){7-8}
 & & \textbf{Acc.} & \textbf{F1} & \textbf{Acc.} & \textbf{F1} & \textbf{Acc.} & \textbf{F1} \\
\midrule
\multirow{2}{*}{Vanilla model} & Mistral-7B & 51.70 & 53.24 & 56.45 & 56.15 & 71.34 & 70.77 \\
 & Llama-2-7B & 72.43 & 83.38 & 61.43 & 64.06 & 62.35 & 69.06 \\
\midrule
\multirow{2}{*}{Distilled model} & Mistral-7B & 65.94 & 69.56 & 67.40 & 67.77 & 83.19 & 84.64 \\
 & Llama-2-7B & 62.08 & 68.24 & 73.98 & 71.15 & 85.42 & 83.45 \\
\midrule
\multirow{2}{*}{Ours} & Mistral-7B & 69.19 & 73.49 & 71.43 & 75.08 & 86.60 & 84.19 \\
 & Llama-2-7B & 82.16 & 87.78 & 74.38 & 77.25 & 87.92 & 86.25 \\
\bottomrule
\end{tabular}
\caption{\small Results on three mental health problem detection tasks. }
\label{tab:generalize}
\end{table*}

\section{Performance Comparison with SOTA Models\footnotemark{}}
This section presents a performance comparison of our proposed method with existing state-of-the-art (SOTA) approaches for mental disorder detection.
\footnotetext{\label{footnote1}We include these sections as a response to reviewers' comments.}
Table~\ref{tab:sota_performance} presents the depression detection performance, comparing our selective distillation framework, applied to student models (Llama-2-7B, Llama-3.1-8B, Mistral-7B), against other competitive SOTA models.
Among these, we particularly highlight the MentalLLaMA models~\cite{yang2024mentallama}, which are specifically fine-tuned for diagnosing mental illness.

As demonstrated in Table~\ref{tab:sota_performance}, our selective distillation method consistently shows strong performance across various student models, significantly surpassing the MentalLLaMA baselines in both accuracy and F1 score. 
These results highlight the effectiveness of our quality-focused approach in enhancing the performance of smaller language models for mental health detection tasks.

\section{Test on Different Domains\footnotemark[\value{footnote}]}
As highlighted in the Conclusion and Limitations sections, our framework, while primarily applied to major depressive disorder detection, possesses inherent versatility across diverse domains.
In this section, we present preliminary experiments demonstrating its applicability and performance improvement across various domains, thereby guiding future work directions.
For this exploration, we utilized three distinct datasets: (1) The Depression Reddit (DR) dataset~\cite{pirina2018identifying}, serving as another major depressive disorder detection task; (2) the Dreaddit dataset~\cite{turcan2019dreaddit}, for stress detection; and (3) the Reddit\_anxiety dataset~\cite{song2023simple}, addressing generalized anxiety disorder detection.
For the stress and anxiety detection tasks, symptom information from the DSM-5 was specifically utilized as the symptom reference.

The experimental results across these three datasets are detailed in Table~\ref{tab:generalize}.
As shown, our method consistently improves both accuracy and F1 score when compared to vanilla and standard distilled models across all datasets.
This consistent enhancement, observed even on out-of-domain datasets like Dreaddit and Reddit\_anxiety, provides strong evidence for the adaptability and broad applicability of our approach.
These findings support our discussion in the Limitations section, indicating that our framework can be effectively adapted to specific domains by leveraging appropriate knowledge sources.



\begin{table*}[t]
\small
    \resizebox{\textwidth}{!}{
        \begin{tabular}{c|p{0.93\textwidth}}
            \hline
            \multicolumn{2}{c}{
                \begin{tabular}[c]{@{}c@{}}
                    Post
                \end{tabular}
                }   \\
            \hline
            \multicolumn{2}{c}{
                \begin{tabular}[c]{@{}p{1\linewidth}@{}}
                    I deal with depression every day. Some days are a complete fog, and I feel like a helpless child. I struggle to hold a job, make enough money, and do things that would make me happy.  My progress feels minimal, and I always end up giving up because my depression depletes nearly all of my energy. I don't know what I'm doing with my life.  I feel so low and beside myself. I have no savings, no plan, no job, and often no will to get up and do things in the morning. My depression makes me feel like a shell of myself. I wouldn't wish this on my worst enemy. I'd give up everything just to not feel this way. I'm still trying, but I'm frustrated and impatient.
                \end{tabular}
            }   \\
            \hline
            Method  & \multicolumn{1}{c}{Responses}    \\
                
            \hline
            +Distillation & \textbf{Yes}. 
            \newline I believe the poster is likely to suffer from very severe Major Depressive Disorder (MDD) for several reasons:
            \newline 1. \textcolor[rgb]{0.8, 0.8, 0.0}{$\triangle$}\textbf{Intensity of negative emotions}: The poster describes feeling ``low'', ``beside myself'', and experiencing a ``fog'' of depression that makes them feel like a ``helpless child''. These phrases suggest a high level of emotional distress.
            \newline 2. \textcolor[rgb]{0.0, 0.5, 0.0}{$\checkmark$}\textbf{Impact on daily life}: The poster reports struggling to hold a job, make enough money, and engage in activities that would bring them joy, indicating a significant impact on their daily life.
            \newline 3. \textcolor[rgb]{0.0, 0.5, 0.0}{$\checkmark$}\textbf{Lack of motivation and interest}: The poster expresses a lack of interest in activities they might enjoy, such as acting or directing, and feels like they have no will to get up and do things in the morning.
            \newline 4. \textcolor[rgb]{0.8, 0.8, 0.0}{$\triangle$}\textbf{Feelings of hopelessness}: The poster states that they wouldn't wish their current state on their worst enemy and would give up all their possessions just to not be depressed, indicating a sense of hopelessness and despair.\\
            \hline
            +Selective & \textbf{Yes}. 
            \newline Here's my step-by-step reasoning:
            \newline 1. \textcolor[rgb]{0.0, 0.5, 0.0}{$\checkmark$}\textbf{Intensity of symptoms}: The poster describes feeling like a ``shell of myself'' and having no energy, which suggests a high level of depressive symptoms.
            \newline 2. \textcolor[rgb]{0.0, 0.5, 0.0}{$\checkmark$}\textbf{Loss of motivation and energy}: The poster mentions that their depression ``depletes nearly all of my energy'', which makes it difficult for them to engage in activities they enjoy or need to do. This is a common symptom of severe depression.
            \newline 3. \textcolor[rgb]{0.8, 0.8, 0.0}{$\triangle$}\textbf{Feelings of hopelessness}: The poster states that they ``don't know what I'm doing with my life'' and feel like they're ``not making any progress''. This sense of hopelessness is a hallmark of severe depression.
            \newline 4. \textcolor[rgb]{0.0, 0.5, 0.0}{$\checkmark$}\textbf{Lack of interest in activities}: The poster mentions that they ``don't want to do anything'' and feel like they're ``just sometimes feel like a shell of myself''. This lack of interest in activities.\\
            \hline
        \end{tabular}
}
\caption{ \small
Example rationales generated by distilled student models assessing a social media post for detecting major depressive disorder and generating a rationale. For these examples, we utilized Llama-3-70B as a teacher model and Mistral as a student model. The notation \textcolor[rgb]{0.0, 0.5, 0.0}{$\checkmark$} indicates the high relevance with clinical symptoms, referring to proper symptoms in the post. 
\textcolor[rgb]{0.8, 0.8, 0.0}{$\triangle$} indicates a sign of depression but not related to diagnostic criteria. To ensure the privacy and anonymity of users, the example post presented in this table have been paraphrased.
}
\label{tab:data:case_study2}
\end{table*}

\end{document}